# Research on Intelligent Aided Diagnosis System of Medical Image Based on Computer Deep Learning

Jiajie Yuan[1], Linxiao Wu[2], Yulu Gong[3], Zhou Yu[4], Ziang Liu[5], Shuyao He[6]
[1]Brandeis University,USA,jiajieyuan@brandeis.edu
[2]Columbia University,USA,wulinxiao1997@gmail.com
[3]Northern Arizona University,USA,yg486@nau.edu
[4]University of Illinois at Chicago,USA,zyu941112@gmail.com
[5] Carnegie Mellon University,USA,ziangliu@alumni.cmu.edu
[6]Northeastern University,USA,he.shuyao@northeastern.edu

*Abstract*—This paper combines Struts and Hibernate two architectures together, using DAO (Data Access Object) to store and access data. Then a set of dual-mode humidity medical image library suitable for deep network is established, and a dual-mode medical image assisted diagnosis method based on the image is proposed. Through the test of various feature extraction methods, the optimal operating characteristic under curve product (AUROC) is 0.9985, the recall rate is 0.9814, and the accuracy is 0.9833. This method can be applied to clinical diagnosis, and it is a practical method. Any outpatient doctor can register quickly through the system, or log in to the platform to upload the image to obtain more accurate images. Through the system, each outpatient physician can quickly register or log in to the platform for image uploading, thus obtaining more accurate images. The segmentation of images can guide doctors in clinical departments. Then the image is analyzed to determine the location and nature of the tumor, so as to make targeted treatment.

## I. INTRODUCTION

Image semantic segmentation has become a research hotspot. At present, the research of machine vision has been paid more and more attention by various research directions. At present, because imaging systems such as CT, MRI, and PET cannot accurately distinguish lesions, the imaging quality of imaging equipment such as CT, MRI, and PET imaging often leads to wrong diagnosis. Therefore, computer-aided imaging means can realize early diagnosis of patients' diseases and improve the curative effect of cancer[1]. This topic intends to carry out research on the segmentation of medical images, focusing on: (1) Effective analysis and recognition of medical images based on target areas in medical images. (2) Applied to 3D reconstruction and display of medical images. (3) Used to determine the size, volume, or volume of a human organ, tissue, or lesion.

The main service object of this system is the radiologist, whose main function is to segment the input image and obtain the corresponding organs or diseased parts, which is convenient for doctors to see, and at the same time, it can also conduct pre-screening[2]. At the same time, the segmentation results of this method can also be used as teaching materials to help medical colleges to train medical talents[3]. By logging in to the account on the platform, and then selecting the type of image to be segmented, the image is finally segmented quickly.

## II. OVERVIEW OF SEGMENTATION MODEL

Because there are many factors such as shadow, background noise and contrast change, how to achieve binary segmentation accurately is a very important problem [4]. If a suitable threshold is selected, the method will misinterpret the target image as noise, and then interpret it as the target, thus reducing the classification accuracy of the image [5]. In recent years, algorithms represented by convolutional neural networks have been applied more and more in many fields. Although convolutional neural networks are effective in automatic image segmentation, it is easy to lose the key boundary during the "pool" operation, which has an adverse effect on the final discrimination [6]. In addition, in practical applications, convolutional neural networks often take a single category marker as the input, but in actual medical imaging systems, it is often necessary to locate the target to a specific area, that is, to assign a category marker to each pixel, which is of great significance for the rapid and efficient completion of biomedical research.

At present, a U-Net algorithm based on the convolutional neural network has been proposed, which adopts 3*3 convolution and drop sampling, and maps the feature information of each subregion by using upper sampling and convolution in the process of decoding, so as to obtain the classification corresponding to each pixel and thus improve the accuracy of the algorithm [7]. However, with the increase of the number of learning layers, the computing resources are more and more occupied, which leads to the overfitting of the model, gradient loss, explosion, and other problems [8]. Through continuous optimization of the U-Net model, this project introduces new ideas such as residual mechanism, dense connectivity mechanism, and double attention mechanism to improve the accurate segmentation and prediction ability of the target region [9]. In this paper, a hybrid algorithm of fusion residual network and U-Net is proposed, and a multi-level prediction algorithm is introduced in the decoding process. Resnet-18 is adopted to replace the traditional U-net structure, and the Plain VGG network is constructed with VGG equation based on the original U-Net architecture [10]. The modeling accuracy is improved by embedding mark mapping in the convolutional network of the Plain VGG. 1) Research on object-oriented multi-level classification and evaluation methods [11]. The U-Net of multi-layer prediction is introduced, and the predicted value of each stage is superimposed with the predicted value of the next stage, thus reducing the training times and improving the prediction accuracy of the model [12]. The algorithm realizes automatic segmentation of diseases such as CT, MRI, and WSI through accurate division of human organs and diseased parts, as well as input of different forms of medical image data, thus achieving the purpose of high precision and high execution speed (Figure 1 is quoted in Sensors, 2021, 21(5): 1794).

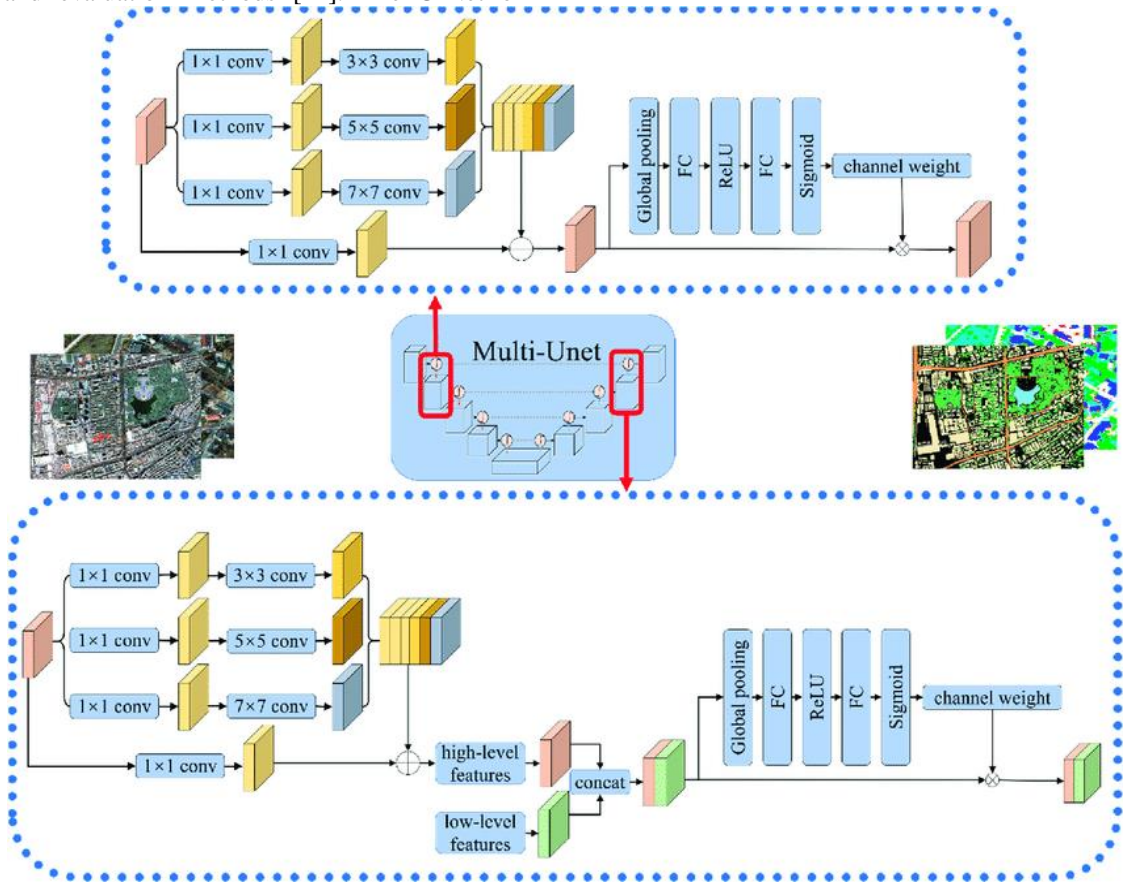

Fig. 1. U-Net structure with multilevel prediction

III. DUAL-MODAL MEDICAL IMAGE-ASSISTED DIAGNOSIS MODEL

The model consists of two parts: a. A feature extraction method based on multi-modal data is proposed. b. Combine the performance of the characteristics [13]. The complete model structure is shown in Figure 2 (image cited in Neurocomputing, Volume 486, 14 May 2022, Pages 135-146).

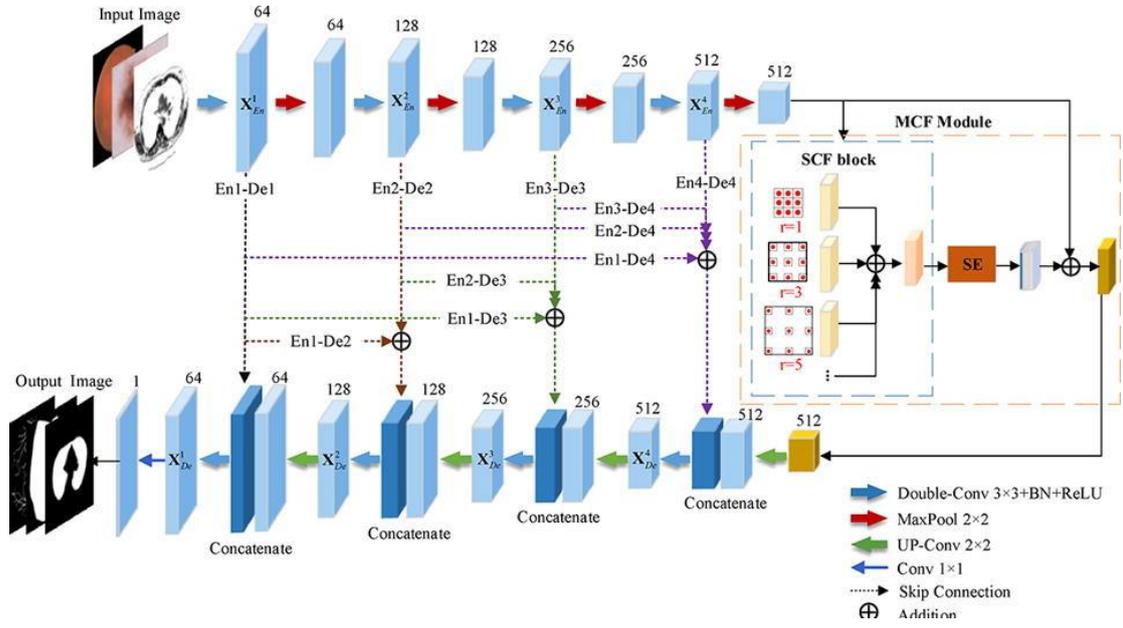

Fig. 2. Dual-modal medical image-assisted diagnosis model

Define dataset $D = \{x_f, x_o \mid f\}$, where $x_f$ and $x_o$ are color fundus images, $x_f, x_o \in R^{224 \times 224 \times 3}$, $f$ is the diagnostic label for this set of images, $f \in$ {neovascular AMD, PCV, others}. The model Wet-AMDNet receives pairs of inputs $\{x_f, x_o\}$ and outputs a diagnosis $\hat{f}$ for the eye.

$$\hat{f} \leftarrow Wet - AMD - Net(\{x_f, x_o\}) \quad (1)$$

The feature-based connection strategy connects $G_f, G_o$ to get vector $G_{con} \in R^{2000}$, and then obtains the score of the final output $\hat{f}$ through the full connection layer.

$$r_{\hat{f},con} = E_{con}G_{con} \quad (2)$$

Where: $E_{con} \in R^{2000 \times 3}, r_{\hat{f},con} \in R^3$. The classification represented in equation (1) is implemented by selecting the category with the highest score.

The feature-based weight allocation strategy obtains the feature vector $G_{add}$ by adding $G_f$ and $G_o$ by weight.

$$G_{add} = \lambda G_f + (1-\lambda)G_o \quad (3)$$

Where: $\lambda$ is a hyperparameter and $0 < \lambda < 1; G_{add} \in R^{1000}$. Then the score of $\hat{f}$ is obtained through the fully connected layer.

$$r_{\hat{f},add} = E_{add}G_{add} \quad (4)$$

Where $E_{add} \in R^{1000 \times 3}, r_{\hat{f},add} \in R^3$. The classification is represented in equation (1)[14]. The weight allocation strategy based on classification results first inputs $G_f$ and $G_o$ respectively into the fully connected layer to get $r_{\hat{f},f}$ and $r_{\hat{f},o}$.

$$r_{\hat{f},f} = E_f G_f; r_{\hat{f},o} = E_o G_o \quad (5)$$

Where: $E_f, E_o \in R^{1000 \times 3}, r_{\hat{f},f} r_{\hat{f},o} \in R^3$. Then add $r_{\hat{f},f}$ and $r_{\hat{f},o}$ by weight is $\hat{f}$.

$$r_{\hat{f},cla} = \lambda r_{\hat{f},f} + (1-\lambda)r_{\hat{f},o} \quad (6)$$

In the formula $r_{\hat{f},cla} \in R^3$, the classification has the highest score.

## IV. REALIZATION OF "INTELLIGENT MEDICAL IMAGE SEGMENTATION SYSTEM"

The whole system uses Python Django library based on MVT architecture. The main line is the interaction between the program and the database [15]. Where V is the view, its role is to show the user the UI interface template, in order to interact with the user; T is for template, that is, the template of the UI interface [16]. This system takes semantic division as the center and uses U-net architecture for semantic division [17]. The overall architecture based on interaction and semantic division is shown in Figure 3 (the picture is quoted in multi-step medical image segmentation based on reinforcement learning).

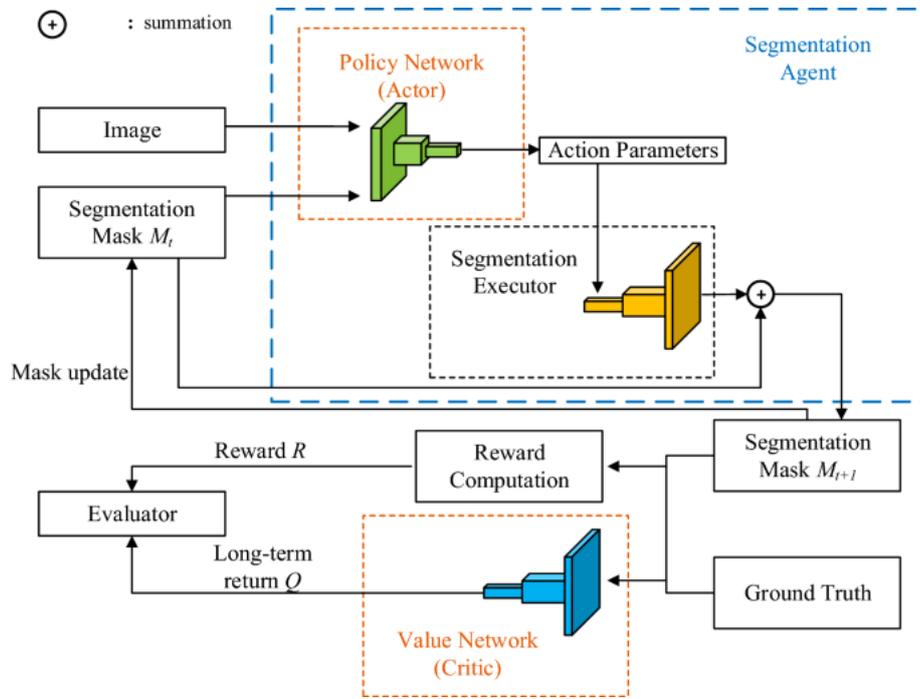

Fig. 3. Structure of intelligent medical image segmentation system

Semantic division refers to the semantic division of images [18]. This framework is a hybrid framework of "compilation and translation", that is, the image is convolved first, then the image is reduced sampling, and then the image is added sampling and convolution processing, so as to classify the image corresponding to each pixel in the image, so as to improve the accuracy of image classification [19]. Because it is image AI processing, it uses NVDIATeslaV100,32 gigabytes of memory and Intel Xeone 5-2650V4 processor 2.2 GHz. In addition, in order to ensure the smooth reading and annotation of doctors, the client memory is set to 8 G, and the bandwidth of a single user is set to 20 M.

A. Implementation of plan

The man-machine interaction module of the system includes login, registration, submission of detection requests, detection result query and download, etc.

*1) Design of Login/Registration View*

In the login View function, the login page is first rendered as the login template and delivered to the client. Then, after receiving the User's login information, through the user mode to see whether there is this user in the table, and can also be verified by the user's name and password, when the verification is passed, you can smoothly switch to the main page. The registration view is also fed back to the client for the first time based on the registration template rendering of the web page, and after the User submits the registration information, the user information is inserted into the table using the user mode.

*2) Implementation of the report check function*

The submit detection function accepts the user's application after the web page is rendered, generates 128 bits of UUID as a task ID of the task, embeds the divided task information into a table, and then inserts the related information of the divided work into a separate task allocation process waiting queue. When the process is finished, the user will be able to perform the following tasks: Switch the full result page of the partition job to the result page of the full partition job of the user.

*3) Design of segmentation effect view*

First, use SegTasks to find all the tasks of users who are currently logging in. Second, by determining the task ID, determine whether the task is waiting, and finally, determine whether the task is finished by determining whether there is a result file. The task assignment thread is completed by the task waiting list, the task waiting queue, and the partition task queue. The task distribution thread is trapped in waiting. When the user asks for a new partition task, inserting it into the task waiting queue will activate the task assignment thread. As for whether this task can be added to the divided task queue, it is necessary to see whether the divided task queue is full at this time, or the first task in the waiting list can be inserted.

*4) Setting of fire indicator Boards*

In the experiment, this paper found that the partition task used alone may conflict with the current work, resulting in execution errors. Therefore, this project proposed a partition method based on security tag bits. The work is added to the queue of the partition, and the thread of execution by the partition adds the new work to the queue of the partition. The task is divided into two parts, that is, the work queue is divided into the security identifier bit, so that it is in the waiting state; Adding a new task to the task list of the partition makes the process in the non-security of layers according to the partition task outline in the model.

*5) Method of image segmentation*

A target classification method based on morphological feature extraction is proposed, and multi-target classification based on classification is carried out, automatic segmentation of multi-modal imaging data based on CT, MRI, etc., and segmentation of pathological images is completed. Firstly, 3D CT and MRI were divided into 2D, and 2D images were segmented and reconstructed successively, and then 3D segmentation of 3D CT and MRI was performed. For pathological images, the sliding window is used to slide, moving the window size by 1/2 at a time, and then calculating the average twice until all the images are moved over to obtain the final segmentation effect. After obtaining the corresponding image, save the image in the corresponding working ID directory and make it in a "safe" state.

## V. SYSTEM APPLICATION EFFECT

(1) On this page, the user can log in directly with the account and password, or register to register the account, log in to the system after registration, and can also perform the action of restoring the password (Figure 4).

(2) When the user logs in successfully, a page will appear, that is, the home page, on this page, the data of the brain tumor, kidney, or kidney tumor to be performed will be entered into the corresponding classification, and then press the "submit" button to segment the image (Figure 5).

(3) When the user submits the task, there will be a partition task list, the list shows the ID, category, submission time, and division of each task, and then the divided picture is classified twice, at that time, the user can have the divided picture displayed in the Image ITK.

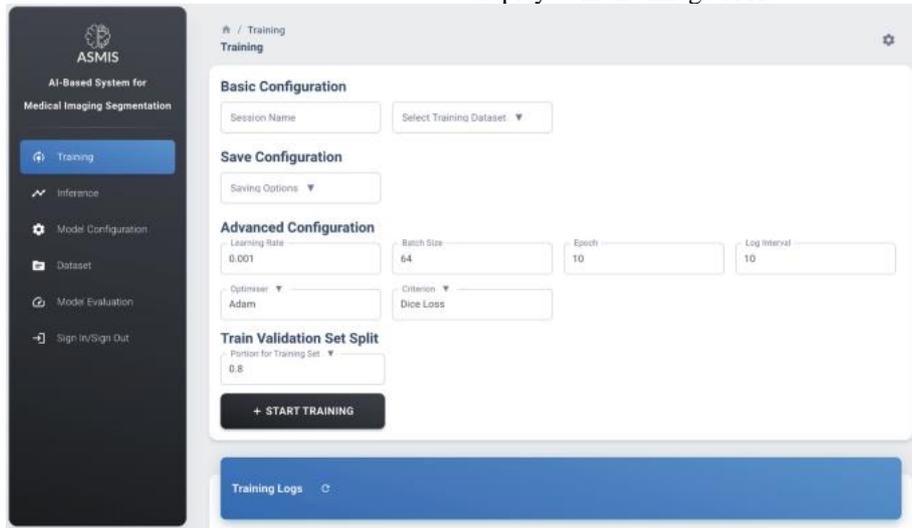

Fig. 4.  System login interface

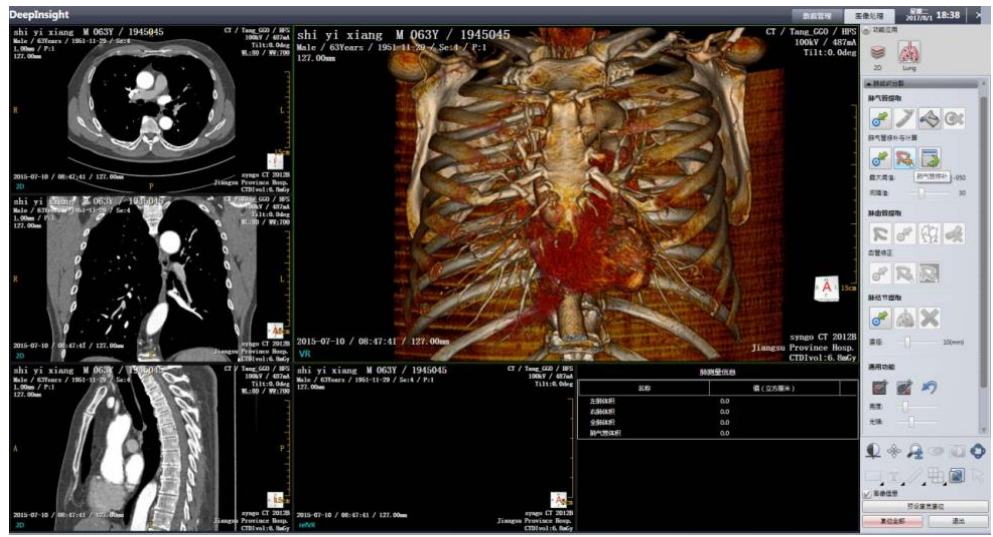

Fig. 5.  Submit task interface

## VI. CLOSING REMARKS

The method adopts MVT architecture and deep neural network based on U-Net transformation as a segmentation model to achieve accurate segmentation and positioning of organs or diseased parts that are difficult to be determined by clinicians, and doctors reduce the burden of medical workers, and improve the accuracy and speed of images. Through the system, each outpatient physician can quickly register or log in to the platform for image uploading, thus obtaining more accurate images. Through image segmentation, doctors in clinical departments can be guided to determine the location and nature of tumors through image analysis, so as to make targeted treatment.